# LDTrack: Dynamic People Tracking by Service Robots using Diffusion Models

Angus Fung · Beno Benhabib · Goldie Nejat

**Abstract**

Tracking of dynamic people in cluttered and crowded human-centered environments is a challenging robotics problem due to the presence of intraclass variations including occlusions, pose deformations, and lighting variations. This paper introduces a novel deep learning architecture, using conditional latent diffusion models, the Latent Diffusion Track (*LDTrack*), for tracking multiple dynamic people under intraclass variations. By uniquely utilizing conditional latent diffusion models to capture temporal person embeddings, our architecture can adapt to appearance changes of people over time. We incorporated a latent feature encoder network which enables the diffusion process to operate within a high-dimensional latent space to allow for the extraction and spatial-temporal refinement of such rich features as person appearance, motion, location, identity, and contextual information. Extensive experiments demonstrate the effectiveness of *LDTrack* over other state-of-the-art tracking methods in cluttered and crowded human-centered environments under intraclass variations. Namely, the results show our method outperforms existing deep learning robotic people tracking methods in both tracking accuracy and tracking precision with statistical significance. Additionally, a comprehensive multi-object tracking comparison study was performed against the state-of-the-art methods in urban environments, demonstrating the generalizability of *LDTrack*. An ablation study was performed to validate the design choices of *LDTrack*.

## 1 Introduction

Service robots need to be able to autonomously track multiple dynamic people in diverse cluttered and crowded human-centered environments to initiate social or physical human-robot interactions (HRI). Such applications range from product searches in retail stores (Dworakowski et al. 2023); directional guidance in airports (Agrawal and Lal 2021) and hospitals (Kollmitz et al. 2019; Vasquez et al. 2017); domestic assistance for elderly and impaired individuals (Sanz et al. 2015); activity/task recognition (T. Liu et al. 2022); robot exploration (Fung et al. 2020; Mohamed et al. 2023; Tan et al. 2023, 2024) and navigation (Gupta et al. 2020; Rebello et al. 2020; Royer et al. 2007; Haitong Wang et al. 2024; Xue et al. 2024); to person following (Pang et al. 2020; Vo et al. 2014; Weber et al. 2018). These environments present the challenge of intraclass variations, which encompass differences in observations of the same people. In particular, in tracking of people, intraclass variations are a result of occlusions, clothing deformations, body articulation, or changes in lighting conditions, which may result in varying appearances for the same individual across different RGB image frames (Fung et al. 2023; Hanxiao Wang et al. 2018; Yan et al. 2023). The intraclass variations in indoor human-centered environments can differ significantly from those in urban outdoor settings. Namely, occlusions in these indoor environments are a result of self-occlusions (people can be in a variety of different poses) or occlusions by non-convex objects such as furniture, leading to disjointed segments of people. In contrast, urban outdoor settings typically include occlusions by convex objects (*e.g.*, vehicles), where the majority of pedestrians are in upright positions, therefore, there are only minor variations in occlusions.

The majority of existing people tracking methods used by robots in human-centered environments require a separate detection system for identification and localization of people and a separate tracking system to track the detected movements of these people over time, while maintaining their identity across different observations (Agrawal and Lal 2021; Kollmitz et al. 2019; Pang et al. 2020; Sanz et al. 2015; Vasquez et al. 2017; Vo et al. 2014; Weber et al. 2018). However, as detection is performed independently on each image frame, these detectors do not have the temporal coherence that is essential for tracking under intraclass variations (Lu et al. 2020). The reliance of these tracking techniques on this isolated frame-by-frame detection can become a bottleneck if the detector fails to identify a person in a subset of the consecutive frames. In contrast, an approach using a single integrated network (joint detection and tracking), rather than having separate detection and tracking modules, is advantageous in terms of computational efficiency for real-time performance of mobile service robots using embedded systems.

A handful of multi-object tracking (MOT) approaches have been developed (Lu et al. 2020; Voigtlaender et al. 2019; Meinhardt et al. 2022; Peize Sun et al. 2021; Zeng et al. 2022; Chuang et al. 2024; Zhou et al. 2020; Chaabane et al. 2021) which incorporate joint detection and tracking (JDT) models to unify these two stages within a single deep learning (DL) network to solve the general object tracking problem. These models utilize proposal boxes that are learned during training to predict the locations of people. However, large datasets are needed to learn these boxes effectively (Dai et al. 2022). For robotics applications with sparse data, this may result in model overfitting, thereby reducing adaptability to diverse and dynamic human-centered settings. Moreover, these JDT methods rely on a fixed number of proposal boxes (Carion et al. 2020) which are for specific sizes and locations of objects including people, this can be challenging to use in crowded environments where people overlap each other significantly. Thus, JDT methods cannot adapt to variations in crowd density, especially when they significantly differ from the training datasets used.

Recently, diffusion models have been proposed for a number of vision-based tasks, including image synthesis (Rombach et al. 2022), inpainting (Lugmayr et al. 2022), object detection (Chen et al. 2023), and object tracking (Luo et al. 2023). Diffusion models are deep generative models which generate data by perturbing an initial input through a forward diffusion process involving Gaussian noise, and a reverse diffusion process using a Markov chain to iteratively recover the original data (Sohl-Dickstein et al. 2015). Diffusion models can be used to generalize tracking across scenes with varying levels of crowd density by iteratively refining proposal boxes generated from noise which do not need to be trained. This is advantageous over the aforementioned learned proposal.

While both (Chen et al. 2023) and (Luo et al. 2023) proposed the use of diffusion models for object detection and object tracking, respectively, these methods use static embeddings that cannot be updated based on information from previous RGB frames. However, in order for diffusion models to address intraclass variations such as partial occlusions, varying lighting conditions and different poses, two key components are necessary. First, it is important to capture temporal person embeddings that can accumulate information and adapt to appearance changes of people over time. Diffusion models have the potential to capture temporal embeddings that are person specific by conditioning on past embeddings from each person's respective tracks, *i.e.* track-conditioned. Secondly, to effectively capture appearance and contextual features over time, the diffusion process needs to use a latent embedding space rather than directly applying the diffusion process over the person bounding box pixel space. To-date, diffusion methods have not yet been explored for the mobile robotic problem of dynamic people tracking.

In this paper, we present a novel people tracking architecture for mobile service robots using conditional latent diffusion models, which we name Latent Diffusion Track (*LDTrack*), to solve the robotic problem of tracking multiple dynamic people under intraclass variations. The main contributions of this paper are: 1) the development of a new people tracking architecture utilizing a joint detection and tracking framework for people tracking by robots in crowded and cluttered environments; 2) the first use of track-conditioned latent diffusion models to capture temporal person embeddings defined as *person track embeddings* – these embeddings are specific to each person and accumulate temporal information of people appearance changes over time; and,

3) the inclusion of a latent feature encoder network to allow for the extraction and spatial-temporal refinement of rich person features including person appearance, motion, location, identity, and contextual information in a high-dimensional latent space.

## 2 Related Work

In this section, we discuss the existing literature on: 1) joint multi-object detection and tracking methods, 2) robotic detection and tracking of people in human-centered environments, and 3) diffusion model methods.

### 2.1 Joint Multi-Object Detection and Tracking Methods

Recent research in multi-object tracking (MOT) utilizes joint detection and tracking (JDT) methods which integrate detection and tracking within a single network (Lu et al. 2020; Voigtlaender et al. 2019; Meinhardt et al. 2022; Peize Sun et al. 2021; Zeng et al. 2022; Chuang et al. 2024; Zhou et al. 2020; Chaabane et al. 2021). While these methods have been designed for tracking general objects, they have mainly been evaluated on urban outdoor datasets with pedestrians and vehicles (Voigtlaender et al. 2019). JDT methods are promising for our robotics problem as a unified network improves temporal coherence as well as computational efficiency (Lu et al. 2020). Both (Lu et al. 2020) and (Voigtlaender et al. 2019) extended off-the-shelf object detectors to perform object tracking. For example, in (Lu et al. 2020), RetinaTrack was developed to improve object tracking efficiency in mission critical tasks such as autonomous driving. RetinaTrack extended the RetinaNet detector (Lin et al. 2020), allowing the detector to extract instance-level object features for data association. RetinaTrack was evaluated on the Waymo Open dataset (Pei Sun et al. 2020), specifically the vehicle class, taken from urban outdoor environments. Similarly, in (Voigtlaender et al. 2019), Track R-CNN (Track Region-based CNN) was developed to jointly perform detection, tracking and segmentation of general objects such as vehicles, with the objective of improving object tracking by considering pixel-level information. Namely, Track R-CNN extended the Mask R-CNN detector (He et al. 2017) by incorporating an associating head for data association and 3D convolution layers for capturing temporal information across frames. Track R-CNN was evaluated on the KITTI MOTS dataset (Voigtlaender et al.

2019) consisting of both vehicles and pedestrians taken from urban outdoor environments.

In (Meinhardt et al. 2022; Peize Sun et al. 2021; Zeng et al. 2022), TrackFormer, TransTrack, and MOTR were introduced as methods using DEtection TRansformers (DETR) (Carion et al. 2020). Their objective was to improve object tracking efficiency. Both TrackFormer (Meinhardt et al. 2022) and MOTR (Zeng et al. 2022) used track queries, which are representations that track objects and maintain their identities across frames. TransTrack (Peize Sun et al. 2021) used object features from previous RGB frames as track queries, and generated parallel sets of detection and tracking boxes which were then combined using the Hungarian algorithm for data association. All three methods were evaluated on the MOT17 (Dendorfer et al. 2021) and MOT20 (Voigtlaender et al. 2019) datasets, consisting of pedestrians in outdoor urban environments.

## 2.2 Robotic Tracking of People

Existing people tracking methods for robotic applications in indoor environments have primarily used tracking-by-detection (TBD) methods to track multiple people (Kollmitz et al. 2019; Munaro and Menegatti 2014; Pang et al. 2020; Pereira et al. 2022; Sanz et al. 2015; S. Sun et al. 2019; Taylor and Riek 2022; Vasquez et al. 2017; Vo et al. 2014; Weber et al. 2018). TBD consist of three main stages (Murray 2017): 1) a static frame-to-frame person detector, 2) prediction of a new person location from previous frames, and 3) data association between frames of predicted and detected person locations. The static frame-to-frame detectors can broadly be divided into: 1) classical methods such as Support Vector Machines (SVMs) with Histogram of Oriented Gradients (HOG) features (Munaro and Menegatti 2014; Sanz et al. 2015; Vo et al. 2014); or 2) deep learning methods such as a) Faster R-CNN (FRCNN) (Kollmitz et al. 2019; Vasquez et al. 2017), b) You Only Look Once (YOLO) (Pang et al. 2020; Pereira et al. 2022; Pinto et al. 2023; Taylor and Riek 2022), and c) Single Shot MultiBox Detector (SSD) (Weber et al. 2018). The prediction stage is used to predict locations of people in future frames using Kalman filters (KF) (Kollmitz et al. 2019; Munaro and Menegatti 2014; Pang et al. 2020; Pereira et al. 2022; Pinto et al. 2023; Sanz et al. 2015; S. Sun et al. 2019; Taylor and Riek 2022; Vo et al. 2014; Weber et al. 2018). The data association stage then matches predictions with new detections using the Hungarian algorithm (Pereira et al. 2022; Vo et al. 2014), the Mahalanobis distance (Kollmitz et al. 2019; Munaro and Menegatti 2014; Pinto et al. 2023; S. Sun et al. 2019), the Euclidean distance (Pereira et al. 2022; Sanz et al. 2015), the Intersection over Union (IoU) distance (Pang et al. 2020; Pereira et al. 2022), or by probability gating (Weber et al. 2018). These methods considered various types of intraclass variations including: partial and/or full occlusions (Kollmitz et al. 2019; Munaro and Menegatti 2014; Pang et al. 2020; Pereira et al. 2022; Pinto et al. 2023; Sanz et al. 2015; S. Sun et al. 2019; Taylor and Riek 2022; Vasquez et al. 2017; Vo et al. 2014; Weber et al. 2018), varying lighting (Vo et al. 2014), and varying poses (Pereira et al. 2022; Vo et al. 2014).

Service robots have used people tracking methods for assistance such as in hospitals and airports (Kollmitz et al. 2019; Taylor and Riek 2022; Vasquez et al. 2017) and in the homes of older adults (Sanz et al. 2015). For example, in both (Kollmitz et al. 2019; Vasquez et al. 2017), a people tracking approach was proposed for robots to find people using mobility aids in populated environments. A FRCNN detector was trained on the Mobility Aids dataset (Vasquez et al. 2017) consisting of RGB-D images of people with different aids in a hospital obtained from a mobile robot. The detector was pretrained on ImageNet (Russakovsky et al. 2015). The position and velocity of a person were then tracked using an extended KF (EKF). For data association between tracks and observations, the pairwise Mahalanobis distances from the EKF were used. Experiments were conducted in an indoor hospital environment and included scenarios with partial person occlusions due to others. In (Taylor and Riek 2022), REGROUP was proposed for assistive robots to guide people in crowded airports. REGROUP detected people using YOLOv3 (Redmon and Farhadi 2018), and performed group data association utilizing KF. Crowd Indication Feature (CIF) was used to determine ego-centric distance metrics to detect crowds. Individual people tracking within crowds was performed by utilizing spatial matrices for precise individual person localizations.

In (Sanz et al. 2015), a people tracking approach was proposed for service robots assisting the elderly and impaired in indoor cluttered environments. A robot used a Kinect RGB-D sensor for depth-based segmentation using Otsu's method to identify potential people and a SVM with HOG features for person detection. For tracking, a KF was used to predict position and velocity of people while a blob matching method based on Euclidean distance was used for data association. Experiments were conducted in an indoor home-like environment and included scenarios with partial person occlusion.

Service robots have also used people tracking methods for person following (Munaro and Menegatti 2014; Pang et al. 2020; Pinto et al. 2023; Weber et al. 2018). For example, in (Pang et al. 2020), a people tracking approach was proposed for a robot following a person in open outdoor environments with a stereo camera. The approach used a YOLO detector for people detection and a KF to predict future locations of people. Data association was performed by matching the detection with the highest IoU score to the predicted boxes from the KF. Experiments were conducted in an outdoor environment with scenarios of partial and full person occlusions.

In (Weber et al. 2018), a people tracking approach was proposed for a robot following a person in unknown crowded environments with an RGB camera. The approach used the SSD (W. Liu et al. 2016) detector for head detection, followed by a KF to predict future people locations. Data association was performed by matching the nearest detection to these predictions based on probability within a specified range. Experiments were conducted in an indoor environment with a robot following a target person who was occluded by others.

In (Munaro and Menegatti 2014), a people tracking approach for a mobile robot used RGB-D data and depth-based sub-clustering to identify and follow individuals. HOG detection was used to detect people's heads within clusters. The method combined multiple clusters together, which could be subdivided due to occlusions, for effective tracking in crowded and occluded environments. An Unscented KF was integrated for estimating the positions and velocities of people. The Mahalanobis distance was used for robust data association. Experiments were conducted in crowded indoor environments under varying lighting conditions.

In (Pinto et al. 2023), a people tracking approach was presented for a companion robot to track, follow and re-identify multiple people. The method detected people in RGB images using YOLOv3 (Redmon and Farhadi 2018), and used TriNet (Yuan et al. 2020) for re-identification and KF for predicting the positions of tracked people. Experiments were conducted in an indoor home environment with occlusions and appearance changes.

Several people tracking methods have also been utilized in general indoor environments (Pereira et al. 2022; S. Sun et al. 2019; Vo et al. 2014). For example, in (Vo et al. 2014), a people tracking approach for a mobile robot was proposed for tracking people under varying poses and appearances. The robot used a Kinect RGB-D sensor for depth-based skin color segmentation which was combined with the Viola-Jones algorithm (Viola and Jones 2001) for face detection, while the upper body detection used an SVM classifier with HOG features extracted from RGB images. The system primarily relied on a fast-compressive tracker, switching to a KF in cases of significant occlusions. For data association, the Hungarian algorithm was used to match detections to existing tracks based on an overlap ratio. Experiments were conducted in an indoor environment under varying lighting conditions with person occlusion.

In (Pereira et al. 2022), two people tracking approaches for mobile robots were compared: 1) SORT (Simple Online and Real-Time Tracking) (Bewley et al. 2016), and 2) Deep-SORT (Wojke et al. 2017). SORT uses a KF for motion prediction, and the Hungarian algorithm for data association, while Deep-SORT extends SORT by integrating a CNN to extract person features. Both methods used the YOLOv3 detector (Redmon and Farhadi 2018) to identify bounding boxes for detected people in RGB images. A new set of data association cost matrices were proposed such as IoU, and Euclidean distances. Experiments were conducted in an indoor campus environments with scenarios consisting of people in sitting and leaning over poses as well as occluding each other.

Lastly, in (S. Sun et al. 2019), an RGB-D multi-person tracking approach was developed for mobile robots in crowded environments. The approach used a deformable part model for person detection and a Gaussian model based on depth values to identify partially occluded people. Data association was performed by matching detections with KF predictions based on Mahalanobis distance. Experiments were conducted on the KTP dataset (Munaro and Menegatti 2014) containing crowds in an indoor office.

## 2.3 Diffusion Models

Diffusion models are deep generative models which operate by gradually transforming a distribution of random noise into a distribution of data (i.e., images or features) through a Markov chain process (Sohl-Dickstein et al. 2015). This process involves a forward diffusion phase, which adds noise to the data, and a reverse diffusion phase, which learns to reverse this process to generate or reconstruct data (Rombach et al. 2022). Diffusion models have been applied to a number of vision tasks, including image synthesis, *e.g.*, (Rombach et al. 2022), inpainting *e.g.*, (Lugmayr et al. 2022), and more recently in object detection (Chen et al. 2023) and object tracking (Luo et al. 2023). Diffusion methods can be conditional and therefore depend on specific conditions to consider more information. For example, conditioning on text descriptions for image generation (Rombach et al. 2022), or conditioning on the visible portion of the image for image inpainting (Lugmayr et al. 2022; Rombach et al. 2022). Diffusion models can also be latent, where the diffusion process occurs in a transformed feature space rather than directly in the pixel space (Rombach et al. 2022).

A conditional latent diffusion model was used in (Rombach et al. 2022) for high-resolution image synthesis by using a sequence of denoising autoencoders. The integration of cross-attention layers enabled the model to handle conditioning inputs such as text descriptions and visible portions of an image. In (Lugmayr et al. 2022), a conditional diffusion model was used for free-form inpainting of images to add missing content to images within regions defined by arbitrary binary masks. The model used a pretrained unconditional diffusion model, conditioning the generation process by sampling unmasked regions, thereby using the visible parts of the image as the condition.

In (Chen et al. 2023), an image-conditioned diffusion model was used for object detection by evolving random boxes from a random distribution to precise detections. During training, the model learned to reverse the diffusion of object boxes from ground truth to randomness. In the inference stage, the model progressively refined randomly generated boxes into predicted bounding boxes, conditioned on the input image. In (Luo et al. 2023), an image-conditioned diffusion model was also used for object tracking by transforming bounding box pairs generated from noise into accurate ground-truth boxes across adjacent video frames. The architecture integrated a YOLOX feature backbone and a data association denoising head. The backbone extracted object features from adjacent frames, which were then refined by the diffusion head through a spatial-temporal fusion module. During training, the model learned to reverse the diffusion of paired object boxes from a random distribution to their ground truth state in pixel space. During inference, the model refined a set of paired randomly generated boxes to object tracks, conditioned on both the current and previous frames.

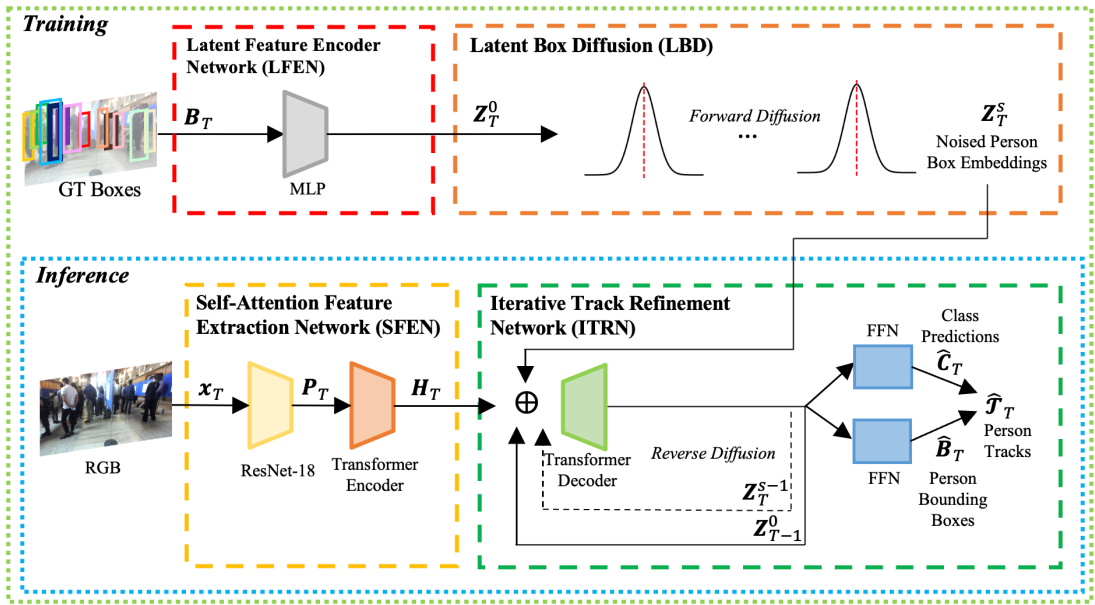

**Fig. 1** Proposed *LDTrack* architecture

### 2.4 Summary of Limitations

As existing robotic people tracking methods are tracking-by-detection methods, they do not consider temporal coherence, as the detector and tracker function independently. The detector processes each frame in isolation, which can lead to inaccuracies in consistently identifying individuals across time, particularly in scenarios with occlusions, lighting changes, and with individuals that have similar appearances. These inaccuracies can result in tracking errors such as identity switches or loss of tracking continuity (Lu et al. 2020). Alternatively, MOT methods (Lu et al. 2020; Voigtlaender et al. 2019; Meinhardt et al. 2022; Peize Sun et al. 2021; Zeng et al. 2022; Chuang et al. 2024; Zhou et al. 2020; Chaabane et al. 2021) have incorporated JDT within a single network architecture. These methods utilize transformer models which use learned proposal boxes for tracking. However, they require large datasets to learn effectively (Dai et al. 2022). Robotic datasets are typically small, leading to potential overfitting. Furthermore, MOT methods are constrained to track from a fixed number of learnable proposal boxes, restricting their ability to adapt to scenes with varying person densities (Chen et al. 2023).

On the other hand, diffusion models have the potential to track people from parameter-free proposal boxes by initializing these boxes from Gaussian noise. Thus, not requiring large datasets for the training of learned proposal boxes as is needed for JDT methods. These boxes can be used to track individuals in environments of any density or complexity, as they can be iteratively refined to track people at any location or of any configuration within the image. To-date, diffusion models have been applied to object detection (Chen et al. 2023) and multi-object tracking (Luo et al. 2023). However, these methods rely on static embeddings that are not updated over time, and do not leverage temporal information over the entire trajectory of a tracked individual (*i.e.*, they do not incorporate dynamic embeddings). As a result, under intraclass variations where an individual is temporarily occluded or not visible in either the previous or current frames, the object tracker may fail to accurately identify the individual.

To address these limitations, we propose *LDTrack*, which utilizes track-conditioned latent diffusion models to accumulate and refine temporal information across an individual's trajectory, allowing *LDTrack* to adapt to changes in appearances over time due to intraclass variations.

## 3 People Tracking Methodology

We propose a novel joint people detection and tracking architecture for tracking multiple dynamic people in cluttered and crowded human-centered environments. The proposed architecture, *LDTrack,* represents the tracking problem as a generative task using conditional latent diffusion models, Fig. 1. Namely, *LDTrack* consists of training and inference sub-systems. The *Inference Subsystem* uses RGB images to determine person feature embeddings which represent person appearance and contextual information via the *Self-Attention Feature Extraction Network* (*SFEN*) module. We use RGB images as they provide the necessary visual features to distinguish between different individuals, for example doctors, nurses, and patients in a hospital. The person *feature* embeddings are utilized by the *Iterative Track Refinement Network* (*ITRN*) module to determine person *track* embeddings. These track embeddings, specific to each individual being tracked, accumulate temporal information by updating over time, capturing person appearance, contextual, motion, location, and identity information. The person track embeddings are used by the two feed-forward neural networks (FFN) for the identification of both individual person bounding boxes and class predictions (person or background). The final output of *LDTrack* is the person tracks, defined as ordered sequences of bounding boxes that represent individual person trajectories within the environment. These tracks are updated based on the identified bounding boxes and class predictions.

The *Training Subsystem* uses the *Latent Feature Encoder Network (LFEN)* module to transform ground truth (GT) person bounding boxes into a latent space representation of higher dimension. This enables the diffusion model to regress richer features such as person appearance, motion, and contextual features in order to maintain accurate identification and tracking of individuals due to changes in their appearances from variations in occlusion, lighting, and poses. The *Latent Box Diffusion* (*LBD*) module introduces Gaussian noise to the latent representations to generate noised person box embeddings, which executes the forward diffusion step of the diffusion model framework, Fig. 1. The noised person box embeddings are then used by the *ITRN* module, which reverse the diffusion process, to update and maintain person tracks. The following subsections discuss the main modules of the inference and training subsystems of *LDTrack* in detail.

### 3.1 Inference Subsystem

The *Inference Subsystem* of the *LDTrack* architecture is used for real-time people tracking in cluttered and crowded human-centered environments. It contains the *SFEN* and *ITRN* modules. The output of the *ITRN* consists of person tracks, representing individual person trajectories within the environment.

#### 3.1.1 Self-Attention Feature Extraction Network (SFEN)

The *SFEN* module extracts person feature (appearance and contextual information) embeddings, $\boldsymbol{H}_T = \{\boldsymbol{h}_T\}, \boldsymbol{h}_T \in \mathbb{R}^D$, from multiple individuals in an RGB image, $\boldsymbol{x}_T \in \mathbb{R}^{H \times W}$, at discrete time $T$. *SFEN* consists of a ResNet network (He et al. 2016), defined as $\mathbf{f}_{CNN}$, and a transformer encoder (Zhu et al. 2021), defined as $\mathbf{f}_{ENC}$. We use a ResNet-18 network to extract initial person features as it is computationally efficient to meet the inference requirements of robotic applications (Fung et al. 2023). Our ResNet-18 network consists of 5 convolution blocks, a fully connected (FC) and a SoftMax layer. The skip connections between the convolutional layers create shortcuts to bypass one or more layers (He et al. 2016). They are used herein to ensure that the network retains low-level details such as color and texture which are crucial for distinguishing between different individuals.

The feature map from the last convolution block of ResNet-18 is flattened into a sequence of $H \times W$ images patches, denoted as $\boldsymbol{P}_T = \{\boldsymbol{p}_T\}$, where each image patch $\boldsymbol{p}_T \in \mathbb{R}^C$ acts as a token, and $C$ is the number of output channels from ResNet-18. The transformer encoder then processes these patch tokens to extract person feature embeddings $\boldsymbol{H}_T = \{\boldsymbol{h}_T\}$. We adapted the transformer architecture from (Zhu et al. 2021), as it utilizes deformable attention modules which allow the encoder to efficiently sample a sparse set of discriminative person features that are most relevant to people tracking under varying environment conditions. The output of the encoder is the person feature embeddings at time $T$:

$$\boldsymbol{H}_T = \mathbf{f}_{SFEN}(\boldsymbol{x}_T) = (\mathbf{f}_{ENC} \circ \mathbf{f}_{CNN})(\boldsymbol{x}_T). \quad (1)$$

#### 3.1.2 Iterative Track Refinement Network (ITRN)

The *ITRN* module consists of a transformer decoder, followed by two FFNs to generate person track embeddings, $\boldsymbol{Z}_T^{s_0} \in \mathbb{R}^{N \times D}$, and corresponding person tracks, $\widehat{\boldsymbol{\mathcal{T}}}_T = \{\widehat{\boldsymbol{T}}_{k_i,T}\}$, where $k_i \in \boldsymbol{\mathcal{K}}_T$ is the subset of people visible in the RGB frame at time $T$. $s_0$ denotes the final step of the reverse diffusion process, where the model has iteratively reconstructed the person track embeddings from its noised state. $\widehat{\boldsymbol{T}}_{k_i,T} = (\widehat{\boldsymbol{b}}_{k_i,T_1}, \widehat{\boldsymbol{b}}_{k_i,T_2}, ..., \widehat{\boldsymbol{b}}_{k_i,T})$ represents the trajectory of a person with identity $k_i$ over timesteps up to the current time $T$, and $\widehat{\boldsymbol{b}}_{k_i,T_j} = (x, y, w, h) \in [0,1]^4$ represents the normalized center coordinates, width, and height of the bounding box. The transformer decoder takes as inputs: 1) the noised person box embeddings, $\boldsymbol{Z}_T^s \in \mathbb{R}^{N_{bx} \times D}$, $\boldsymbol{Z}_T^s \sim \mathcal{N}(\boldsymbol{0}, \boldsymbol{I})$, initialized from a multivariate normal distribution, where $N_{bx}$ represents the number of box embeddings, 2) person track embeddings $\boldsymbol{Z}_{T-1}^{s_0} \in \mathbb{R}^{N_{tk} \times D}$ from the previous timestep, where $N_{tk}$ represents the number of track embeddings from $T-1$, and 3) per-son feature embeddings $H_T$ from the *SFEN* module. The decoder produces a set of output person track embeddings $Z_T^{s_0}$ of size $N = N_{\text{bx}} + N_{\text{tk}}$. The noised person box embeddings $Z_T^{s}$ are the starting point for the reverse diffusion process, which systematically denoises and refines these embeddings into the final set of output person box embeddings $Z_T^{s_0}$. Person box embeddings represent probabilistic proposals for potential person locations (position and scale) within a frame.

The structure of the transformer decoder is an extension of the decoder in (Zhu et al. 2021). It incorporates six decoder layers with each layer containing a combination of multi-head attention mechanisms and FFNs. Fig. 2 presents the tokens and embeddings used in the transformer encoder and decoder networks. The input tokens to the encoder are the $H \times W$ image patches $P_T$, and the corresponding output are the person feature embeddings $H_T$, both represented by grey squares. In the decoder, the $N_{\text{bx}}$ noised person box embeddings $Z_T^{s}$ (white, squares, Fig. 2) and the $N_{\text{tk}}$ person track embeddings $Z_{T-1}^{s_0}$ from the previous timestep (colored squares, Fig. 2) are combined together as the input. Cross-attention is applied between $Z_T^{s}$ and $Z_{T-1}^{s_0}$, and the image patches $P_T$. The decoder performs iterative refinement, denoising the person box and track embeddings with each iteration.

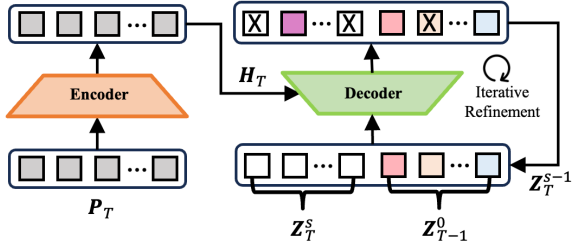

**Fig. 2** Tokens and embeddings used in the *SFEN* and *ITRN* modules. Grey squares represent patch tokens, white squares represent noised person box embeddings, and colored squares represent person track embeddings. Squares with "X" represent embeddings corresponding to inactive person tracks.

Unique to our approach, we utilize noised box embeddings $Z_T^{s}$ and track embeddings from the previous timestep $Z_{T-1}^{s_0}$ as the inputs instead of learned position embeddings which were used in (Zhu et al. 2021). The transformer decoder performs the reverse diffusion process, with $Z_T^{S}$ and $Z_{T-1}^{s_0}$ concatenated at the input. The inputs then undergo cross-attention with the person feature embeddings $H_T$ from *SFEN* to produce person track embeddings $Z_T^{0}$, allowing the decoder to focus on specific regions of an RGB image. These embeddings can be refined iteratively by reversing the diffusion steps in a sequential manner, starting from $s = S$ until reaching the step $s = s_0 = 0$ through the $\mathbf{f}_{\text{ITRN}}$ network:

$$Z_T^{S-1} = \mathbf{f}_{\text{ITRN}}\left(Z_T^{S}, Z_{T-1}^{s_0}; H_T\right), \qquad (2)$$

where the person track embeddings $Z_T^{S-1}$ at diffusion step $S - 1$

are obtained from step $S$. The iterative refinement process allows for incremental improvements of tracking in latent space, enabling *LDTrack* to adapt to changes in individual appearances as well as accommodate diverse spatial configurations of varying crowd densities. Thus, these track embeddings are refined both spatially and temporally to learn and accumulate temporal person features that are robust under intraclass variations. Spatial refinement is achieved through the aforementioned iterative process of reversing the diffusion steps, and temporal refinement is achieved by conditioning on the person track embeddings from previous timesteps.

After iterative person track refinement, the final set of output person track embeddings $Z_T^{0} \in \mathbb{R}^{N \times D}$, are then passed into two separate FFNs. One FFN is used for person class prediction $\hat{c}_{k_i,T}$ (person or background), and the other for person bounding box prediction $\hat{b}_{k_i,T}$. The final output $\hat{Y}_T$ of the FFN consists of the predicted person bounding box and class pairs:

$$\hat{Y}_T = \{\hat{y}_{k_i,T}\}_{i=1}^{N_{\text{bx}}+N_{\text{tk}}}, \mathbf{y}_{k_i,T} = (\hat{c}_{k_i,T}, \hat{b}_{k_i,T}). \qquad (3)$$

Person tracks, which are ordered sequences of bounding boxes representing a person's trajectory and their respective person track embeddings, can become inactive by no longer being updated or tracked. Namely, they become inactive when the individual is no longer visible in an RGB image, for example, due to full occlusion. A track becomes inactive (squares with "x" in Fig. 2): if $p(\hat{c}_{k,T}) < \sigma_{\text{cls}} = 0.4$ or if IoU $< \sigma_{\text{IoU}} = 0.9$ with other bounding boxes (Meinhardt et al. 2022). New person tracks $\hat{T}_{k_i,T}, k_i \in \mathcal{K}_{\text{new}}$ are obtained from $Z_T^{s}$ for new detections, while existing person tracks $\hat{T}_{k_i,T}, k_i \in \mathcal{K}_{T-1}$ are updated from $Z_{T-1}^{s_0}$. A person track $\hat{T}_{k_i,T}$ can be obtained using the predicted person bounding boxes $\hat{b}_{k_i,T}$ from Eq. (3) as follows:

$$\hat{T}_{k_i,T} = \begin{cases} \hat{T}_{k_i,T-1} \cup \{\hat{b}_{k_i,T}\}, & k_i \in \mathcal{K}_{T-1} \\ \{\hat{b}_{k_i,T}\}, & k_i \in \mathcal{K}_{\text{new}} \end{cases}. \qquad (4)$$

Thus, the number of tracks $N_{\text{tk}}$ changes when new tracks are created or existing tracks become inactive.

To handle temporary occlusions, where a person may be obscured from the view of the robot's RGB camera for a short period of time, inactive person tracks are retained for an additional $N_{\text{reid}}$ frames similar to (Meinhardt et al. 2022). Should the person reappear within this short time span, *LDTrack* can resume tracking without initializing a new person track for an existing person. These person tracks can be reactivated if their classification score $p(\hat{c}_{k,T}) > \sigma_{\text{cls}}$, indicating a renewed confidence in the presence of a person (Meinhardt et al. 2022). Since person track embeddings from prior timesteps undergo direct refinements, data association is implicitly achieved without any additional computational overhead. Thus, the final set of person tracks $\hat{\mathcal{T}}_T$ obtained from the inference subsystem is:

$$\hat{\mathcal{T}}_T = \{\hat{T}_{k_i,T}\}, k_i \in \mathcal{K}_T = \mathcal{K}_{T-1} \cup \mathcal{K}_{\text{new}}. \qquad (5)$$

## 3.2 Training Subsystem

The *Training Subsystem* uses both the *LFEN* and *LBD* modules to produce noised person box embeddings $\boldsymbol{Z}_T^S$. The *ITRN* then transforms these noised person box embeddings $\boldsymbol{Z}_T^S$ into person bounding box predictions, $\widehat{\boldsymbol{Y}}_T$, to optimize a set prediction loss.

### 3.2.1 Latent Feature Encoder Network (LFEN)

The *LFEN* module converts ground truth (GT) person bounding boxes into a higher dimensional latent space. $\boldsymbol{B}_T \in \boldsymbol{\mathcal{B}}$ is the set of GT person bounding boxes at discrete time $T$, and $\boldsymbol{b}_{k,T} \in \boldsymbol{B}_T$ represent the person bounding box with identity $k \in \boldsymbol{\mathcal{K}}$, where $\boldsymbol{b}_{k,T} \in [0,1]^4$ as defined previously. The latent representation of the GT person bounding boxes $\boldsymbol{z}_{k,T} \in \mathbb{R}^D$ are obtained by passing the set of GT bounding boxes $\{\boldsymbol{b}_{k,T}\}$ through the multi-layer perception (MLP), $\mathbf{f}_{\text{LFEN}}$:

$$\boldsymbol{z}_{k,T} = \mathbf{f}_{\text{LFEN}}\left(\boldsymbol{b}_{k,T}\right). \tag{6}$$

We designed the MLP to consist of three fully connected layers, each with a hidden layer dimension of size $D$. To account for the varying number of ground truth boxes $\boldsymbol{b}_{k,T}$ across different RGB frames and also maintain consistent input dimensions for the *LBD* module, we pad $\boldsymbol{B}_T$ to the fixed size $N_{\text{bx}}$, which is the same size as the noised person box embeddings (Chen et al. 2023). This is achieved by initializing these GT bounding boxes from a normal distribution:

$$\boldsymbol{b}_{\text{pad}} \sim \mathcal{N}(0.5, \sigma^2), \tag{7}$$

where $\sigma = 1/6$ was chosen such that $3\sigma$ represents a range between 0 and 1. The output set of latent representations of person box embeddings $\boldsymbol{Z}_T = \{\boldsymbol{z}_{k,T}\}_{i=1}^{N_{\text{bx}}}$ is then provided to the *LBD* module.

### 3.2.2 Latent Box Diffusion (LBD)

The objective of the *LBD* module is to initiate the forward diffusion process by converting the set of latent representations of person box embeddings $\boldsymbol{Z}_T$ to noised person box embeddings $\boldsymbol{Z}_T^S$. This is achieved through a Markovian-chain driven diffusion forward process by systematically applying Gaussian noise to the data according to the noise variance schedule $\beta_1, \ldots, \beta_S$ (Ho et al. 2020):

$$q(\boldsymbol{Z}_T^s|\boldsymbol{Z}_T^0) = \mathcal{N}(\boldsymbol{Z}_T^s|\sqrt{\overline{\alpha_t}}\boldsymbol{Z}_T^0, (1 - \overline{\alpha_s})\boldsymbol{I}, \tag{8}$$

where $q(\boldsymbol{Z}_T^s|\boldsymbol{Z}_T^0)$ is the forward diffusion noise distribution of the person box embeddings $\boldsymbol{Z}_T^s$ at diffusion step $s$ given the initial person box embeddings; $\boldsymbol{Z}_T = \boldsymbol{Z}_T^0 = \{\boldsymbol{z}_{k,T}^0\}$ at step $s = 0$; $\alpha_s = 1 - \beta_s$; $\overline{\alpha_s} = \prod_{j=1}^{s} \alpha_j$; and $\boldsymbol{I}$ is the identity matrix. $\alpha_s$ represents the proportion of the original embedding that is retained after noise has been applied and $\beta_s$ represents the amount of noise being added at each step $s$ in the diffusion process. The output set of noised person box embeddings, at the final step $s =$ $S$, is $\boldsymbol{Z}_T^S = \{\boldsymbol{z}_{k,T}^S\}$. This output set is provided to the *ITRN* module.

### 3.2.3 Self-Attention Feature Extraction Network (SFEN)

The SFEN module is trained on pairs of consecutive RGB frames $(\boldsymbol{x}_{T-1}, \boldsymbol{x}_T)$ in order to learn temporal features of people as their appearance changes across RGB frames. The pair of frames is used in Eq. (1) to extract the corresponding person feature embeddings $\boldsymbol{H}_{T-1}$ and $\boldsymbol{H}_T$. The person feature embeddings are then also provided to the *ITRN* module.

### 3.2.4 Iterative Track Refinement Network (ITRN)

The *ITRN* uses $\boldsymbol{H}_{T-1}$ and $\boldsymbol{H}_T$ to predict person class and bounding boxes which are used to optimize a set prediction loss. For $\boldsymbol{x}_{T-1}$, we use $\mathbf{f}_{\text{ITRN}}$ to directly predict the person box embeddings $\boldsymbol{Z}_{T-1}^{s_0}$ within a single step of reverse diffusion using Eq. (2):

$$\boldsymbol{Z}_{T-1}^{s_0} = \mathbf{f}_{\text{ITRN}}\left(\boldsymbol{Z}_{T-1}^S; \boldsymbol{H}_{T-1}\right). \tag{9}$$

Namely, time $T - 1$ serves as the starting point for each RGB pair. As discussed in Section 3.1.2, the person box embeddings $\boldsymbol{Z}_{T-1}^{s_0}$ are passed into two separate FFNs for person bounding boxes $\widehat{\boldsymbol{b}}_{k_i,T}$ and class predictions $\widehat{c}_{k_i,T}$. $\boldsymbol{Z}_{T-1}^{s_0}$ which satisfy the IoU and classification thresholds $\sigma_{\text{IoU}}$ and $\sigma_{\text{cls}}$ are retained as person track embeddings for the subsequent timestep. At $T$, $\mathbf{f}_{\text{ITRN}}$ is again used to directly predict person box embeddings $\boldsymbol{Z}_T^{s_0}$ within a single step from $\boldsymbol{Z}_T^{s_0}$ using Eq. (2):

$$\boldsymbol{Z}_T^{s_0} = \mathbf{f}_{\text{ITRN}}\left(\boldsymbol{Z}_T^S, \boldsymbol{Z}_{T-1}^{s_0}; \boldsymbol{H}_T\right). \tag{10}$$

$\boldsymbol{Z}_T^{s_0}$ are also passed into the FFNs to obtain $\widehat{\boldsymbol{b}}_{k_i,T}$ and $\widehat{c}_{k_i,T}$, respectively. Thus, the final output $\widehat{\boldsymbol{Y}}_T$ of the FFNs consists of the predicted person bounding box and class pairs, Eq. (3). We use the set prediction loss (Carion et al. 2020) on $\widehat{\boldsymbol{Y}}_{T^*}$. To compute the loss, a bipartite matching between the predictions $\widehat{\boldsymbol{Y}}_T$ and ground truth $\boldsymbol{Y}_T$ is required, Fig. 3. We adopted a matching strategy similar to (Meinhardt et al. 2022), where we divide the ground truth $\boldsymbol{Y}_T = \{\boldsymbol{y}_{k_i,T}\}_{i=1}^M$ into two subsets based on their identities: 1) $\boldsymbol{Y}_{1,T} = \{\boldsymbol{y}_{k_i,T} \mid k_i \in \boldsymbol{\mathcal{K}}_T \cap \boldsymbol{\mathcal{K}}_{T-1}\}$, which represents people who have already been tracked in frame $\boldsymbol{x}_{T-1}$ and present in the current frame $\boldsymbol{x}_T$; and 2) $\boldsymbol{Y}_{2,T} = \{\boldsymbol{y}_{k_i,T} \mid k_i \in \boldsymbol{\mathcal{K}}_T \backslash \boldsymbol{\mathcal{K}}_{T-1}\}$, which represents new people present in the current frame $\boldsymbol{x}_T$. Both of these subsets $\boldsymbol{Y}_{1,T}$ and $\boldsymbol{Y}_{2,T}$ are labeled in Fig. 3. The ground truth $\boldsymbol{Y}_T$ is padded with $\emptyset$ ($c_i = 0$, background class) such that it has the same size as the unmatched pairs, namely $(N_{\text{bx}} + N_{\text{tk}}) - |\boldsymbol{\mathcal{K}}_T \cap \boldsymbol{\mathcal{K}}_{T-1}|$ (white circles, Fig. 3). For subset 1, we match the ground truth $\boldsymbol{y}_{k_i,T}$ to the prediction $\widehat{\boldsymbol{y}}_{k_i,T}$ with the same identity (*i.e.*, same color in Fig. 3), as person track embeddings from $T - 1$ carry over their identity to the subsequent frame $T$. In subset 2, we match the ground truth and remaining predictions (grey circles, Fig. 3) based on an injective minimum cost mapping in order to find the permutation $\tau$ with the lowest cost (Carion et al. 2020):

$$\hat{\tau} = \arg\min_\tau \sum_{y_i \in Y_{2,T}} \mathcal{C}_{\text{match}}(y_i, \hat{y}_{\tau(i)}), \quad (11)$$

where $\mathcal{C}_{\text{match}}(y_i, \hat{y}_{\tau(i)})$ is the pairwise matching cost between the ground truth $y_i$ and prediction $\hat{y}_{\tau(i)}$ with index $\tau(i)$. This matching cost is defined as:

$$\mathcal{C}_{\text{match}}(y_i, \hat{y}_{\tau(i)}) = \lambda_{\text{cls}} \hat{p}_{\tau(i)}(\hat{c}_i) + \lambda_{l1} |b_i - \hat{b}_{\tau(i)}| + \lambda_{\text{gIoU}} \mathcal{C}_{\text{gIoU}}(b_i, \hat{b}_{\tau(i)}), \quad (12)$$

where $\lambda_{\text{cls}}$, $\lambda_{l1}$, and $\lambda_{\text{gIoU}}$ are weighting parameters, and $\mathcal{C}_{\text{gIoU}}$ is the generalized IoU loss (Rezatofighi et al. 2019). The optimal assignment of the ground truth subset $Y_2$ is obtained with the Hungarian algorithm, which uses a cost matrix based on the matching cost (Kuhn 1955).

After obtaining the optimal assignments, the final loss, $\mathcal{L}_{\text{MOT}}$, is computed. $\mathcal{L}_{\text{MOT}}$ is the set prediction loss computed over all the output person bounding box predictions $\hat{Y}_T = \{\hat{y}_{k_i,T}\}_{i=1}^{N_{\text{bx}}+N_{\text{tk}}}$:

$$\mathcal{L}_{\text{MOT}}(y_i, \hat{y}_{\tau(i)}) = \sum_{y_i \in Y_t} \lambda_{\text{cls}} \hat{p}_{\tau(i)}(\hat{c}_i) + \left(\lambda_{l1} |b_i - \hat{b}_{\tau(i)}| + \lambda_{\text{gIoU}} \mathcal{L}_{\text{gIoU}}(b_i, \hat{b}_{\tau(i)})\right) \mathbf{1}_{c_i=1}. \quad (13)$$

The objective of the *Training Subsystem* is to minimize $\mathcal{L}_{\text{MOT}}$ such that the predicted person bounding boxes and classes $Y_T$ from *LDTrack* closely align with the ground truth $\hat{Y}_T$.

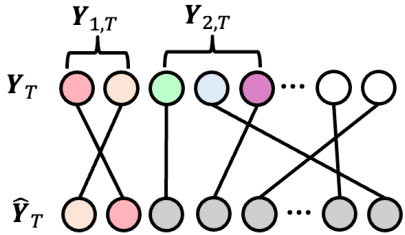

**Fig. 3** Bipartite matching between the ground truth $Y_T$ and predictions $\hat{Y}_T$, where the colored circles in the $\hat{Y}_T$ and $Y_T$ rows represent boxes corresponding to existing person tracks and ground truth person tracks, respectively; the grey circles represent no-identity boxes, and the white circles represent padded boxes.

## 4 Datasets

We utilized multiple datasets to train and evaluate our people detection and tracking *LDTrack* architecture to include a variety of intraclass variations and environmental conditions. The following datasets were used for training:

**1. The InOutDoor (IOD) Dataset** (Mees et al. 2016) consisting of a total of 8,300 RGB-D images, each with a resolution of 950x540 pixels at 30 Hz. The dataset contains four image sequences of people, including groups of up to six people, walking under varying lighting conditions. Images were obtained from a Microsoft Kinect sensor mounted on the top of a mobile robot in both an indoor and outdoor campus environment. The first three sequences were used for training, and the last sequence was used for testing.

**2. Kinect Tracking Precision (KTP) Dataset** (Munaro and Menegatti 2014) consists of 8,475 RGB images with a resolution of 640x480 pixels at 30 Hz. The images are obtained from a side-mounted Microsoft Kinect sensor on a mobile robot in an indoor office environment. The dataset contains four RGB sequences representing arc, still, rotation, and translation robot motion patterns. A total of 14,766 instances of people are in the dataset, with groups as large as five individuals. The individuals are often partially occluded by others and self-occluded.

**3. ISR Tracking (ISRT) Dataset** (Pereira et al. 2022) consists of 10,000 RGB images with a resolution of 640x480 pixels at 30 Hz. The images are obtained from an Intel RealSense D435 sensor mounted on a mobile robot in an indoor university campus setting. The dataset includes annotations for ten different object classes, including a person class. For training, we only considered the labels corresponding to people. We used odd-numbered sequences for training, and even-numbered sequences for testing (Pereira et al. 2022). However, for testing, we specifically selected a subset of sequences that contain people (*e.g.*, 4, 6, and 12). The dataset contains groups of up to five people with scenarios of people occluding each other, body part occlusions, objects occluding individuals, and diverse poses such as sitting, leaning and standing.

**4. Robots in Crowded People Conferences (R-CPC) Dataset:** We collected our own R-CPC dataset from three different indoor university campus conferences using a mobile robot developed in our lab. A ZED Mini camera was mounted on each side of the robot (four in total), Fig. 4. The R-CPC dataset consists of 2,575 RGB-D images from each camera, all with a resolution of 640x480 pixels at 30 Hz. The Train Set 1 and Test Set 1 include scenarios of the robot guiding a crowd of people from a main lobby down a corridor in a building. Set 1 consists of a total of

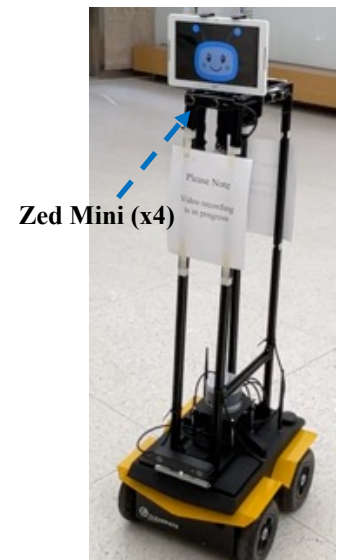

**Zed Mini (x4)**

**Fig. 4** Mobile robot used to collect the R-CPC Dataset.

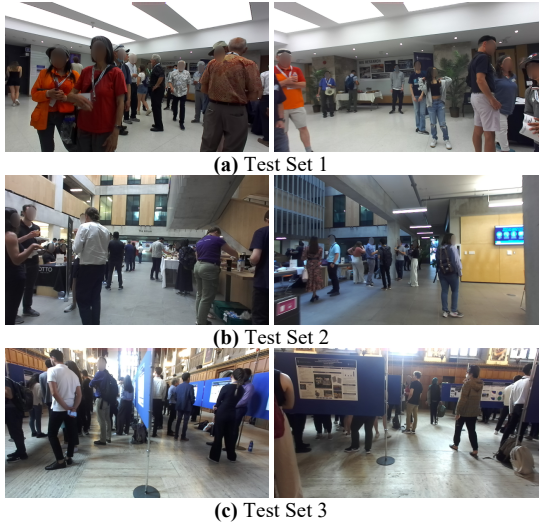

**(a)** Test Set 1

**(b)** Test Set 2

**(c)** Test Set 3

**Fig. 5** Example Images from the R-CPC Dataset.

408 (for training) and 277 (for testing) images with 3,368 and 1,452 instances of people, respectively. The Train Set 2 and Test Set 2 include the robot moving within crowds of people that are walking, standing, and sitting at a conference. Set 2 consists of a total of 370 (for training) and 331 images (for testing) with 2,738 and 2,294 instances of people. Lastly, the Train Set 3 and Test Set 3 include scenarios of a robot guiding crowds of people from the main entrance of a building to a large poster room, where the robot also moves among the presenting attendees. Set 3 consists of a total of 808 (for training) and 380 (for testing) images with 4,713 and 2,603 instances of people. Example images from the R-CPC dataset is presented in Fig. 5.

The R-CPC dataset uniquely contains dense crowds of up to 18 people per frame. It also includes: 1) self-occlusions, partial occlusions, and full occlusions in both large open areas and within narrow corridors; 2) frequent short-term full occlusions as individuals momentarily walk in front of the camera; 3) frequent crossings where paths of multiple people intersect; 4) motion blurs due to rapid movements of individuals; 5) large and small objects occluding individuals (*e.g.*, tables and chairs) and conference equipment (*e.g.*, laptops, monitors); 6) pose variations (*e.g.*, standing, bending over, sitting, and clothing deformations (*e.g.*, jackets, backpacks, shoulder bags); and 7) a wide range of lighting conditions, from dim to bright environments due to the presence of both natural and artificial lighting sources.

## 5 Training

*LDTrack* was trained using the *Training Subsystem* discussed in Section 3.2. We used a batch size of 1, and learning rates for the CNN and transformer network of 0.00001 and 0.0001, respectively (Carion et al. 2020). The weighting parameters $\lambda_{cls}$, $\lambda_{l1}$, and $\lambda_{gIoU}$ were set to 2, 5, and 2 (Carion et al. 2020). For the

MLP in the *LFEN* module, we used a latent dimension $D = 288$. For the *LBD* module, we select $N_{bx} = 500$. We selected parameters as it provides the optimal balance between detection accuracy and number of refinement steps for the reverse diffusion process (Chen et al. 2023). *LDTrack* was trained and validated separately on each of the four datasets with early stopping. Training took place on a workstation with two RTX 3070 GPUs, an AMD Ryzen Threadripper 3960X, and 128GB of memory.

## 6 Experiments

The objective of the experiments was to investigate the tracking accuracy and precision of *LDTrack* under intraclass variations. Specifically, 1) we compare our *LDTrack* with state-of-the-art (SOTA) robot people tracking methods, 2) we compare our *LDTrack* with SOTA multi-object tracking methods to evaluate the efficacy of diffusion models for tracking pedestrians in urban environments, and 3) we conduct an ablation study to validate the design choices of our architecture.

### 6.1 Person Tracking Experiments

We conduct a comparison study to evaluate the performance of *LDTrack* with SOTA robot person tracking methods in environments with varying levels of person occlusion, illumination, and deformation.

#### 6.1.1 Performance Metrics

We used the CLEAR MOT (Bernardin and Stiefelhagen 2008) evaluation metrics, a standard evaluation tool which have been designed for assessing object tracking performance. The CLEAR MOT metrics we use herein are:

1. **Multiple Object Tracking Precision (MOTP)**: MOTP measures the average overlap between the ground truth and predicted people bounding boxes:

$$\sum_{i=1}^{N_{match}} \text{IoU}\left(\boldsymbol{b}_i, \hat{\boldsymbol{b}}_i\right)/N_{match}, \qquad (14)$$

where $N_{match}$ is the total number of matches between the ground truth and prediction. A higher MOTP value, up to 100, indicates that the predicted person bounding boxes more closely match the actual person locations.

2. **Multiple Object Tracking Accuracy (MOTA):** MOTA measures tracking performance by taking into account the number of: false positives $N_{fp}$, false negatives $N_{fn}$, identity switches $N_{IDSW}$ (when two tracked individuals mistakenly swap their identities), and ground truth person boxes $N_{GT}$:

$$\text{MOTA} = 1 - (N_{FP} + N_{FN} + N_{IDSW})/N_{GT}. \qquad (15)$$

A MOTA value close to 100 indicates high tracking accuracy.

#### 6.1.2 Comparison Methods

**Table 1** Tracking Accuracy Comparison of our Proposed Tracking Method versus Existing Tracking Methods

| Dataset / Method | IOD | | KTP | | ISRT | | R-CPC | | | | | |
|---|---|---|---|---|---|---|---|---|---|---|---|---|
| | Test Set Lighting | | Test Set Occlusion | | Test Set Occlusion/Pose | | Test Set 1 All | | Test Set 2 All | | Test Set 3 All | |
| | MOTP | MOTA | MOTP | MOTA | MOTP | MOTA | MOTP | MOTA | MOTP | MOTA | MOTP | MOTA |
| *LDTrack* | 83.1 | 78.6 | 93.1 | 92.7 | 82.8 | 80.9 | 83.9 | 49.0 | 80.6 | 43.9 | 81.6 | 57.7 |
| SORT (Pereira et al. 2022) | 70.0 | 57.3 | 65.6 | 71.6 | 81.4 | 69.0 | 74.4 | 30.6 | 72.9 | 24.0 | 73.2 | 36.4 |
| DeepSORT (Pereira et al. 2022) | 70.1 | 54.3 | 64.6 | 62.8 | 81.3 | 67.6 | 72.3 | 35.6 | 71.7 | 29.4 | 71.4 | 38.5 |
| STARK (Pinto et al. 2023) | 70.1 | 52.5 | 64.5 | 63.3 | 81.2 | 67.1 | 72.1 | 29.5 | 71.0 | 23.2 | 71.4 | 35.6 |
| REGROUP (Taylor and Riek 2022) | 71.0 | 32.0 | 64.6 | 57.3 | 81.2 | 38.0 | 75.3 | 38.4 | 67.1 | 32.9 | 74.9 | 41.2 |
| Munaro et al. (Munaro and Menegatti 2014) | - | - | 84.2 | 86.1 | - | - | - | - | - | - | - | - |
| Sun et al. (S. Sun et al. 2019) | - | - | 84.4 | 88.2 | - | - | - | - | - | - | - | - |

Note: all missing values above (denoted '-') were not reported.

We compared *LDTrack* with the following RGB SOTA people tracking methods for mobile robots:

**1. SORT and DeepSORT** (Pereira et al. 2022): SORT and DeepSORT both use the YOLOv3 detector to identify bounding boxes of objects in RGB images. KF is used for motion prediction and the Hungarian algorithm is used for data association. These two methods were selected as they have achieved SOTA results in tracking people under occlusions, with the highest people tracking accuracy and precision on the ISRT dataset, designed for assistive mobile robots (Pereira et al. 2022). However, it should be noted that while (Pereira et al. 2022) reported results for general object tracking across nine different categories, for our comparison, we specifically used only the video sequences containing people as discussed in Section 3.

**2. REGROUP** (Taylor and Riek 2022): REGROUP also uses YOLOv3 to detect people in RGB images, a KF for motion prediction, and a CNN for appearance-based data association. REGROUP was selected as its Crowd Indication Feature can provide accurate discrimination of individual spatial positions within crowds of people (Taylor and Riek 2022). This method is considered SOTA for tracking people within crowds as it achieved the highest results in tracking people in an environment with varying lighting conditions and crowded scenes. Furthermore, REGROUP has shown to have real-time capabilities and is robust to occlusion, camera egomotion, shadows, and varying lighting conditions (Taylor and Riek 2022).

**3. STARK** (Pinto et al. 2023): STARK uses YOLOv3 for detection and KF for motion prediction. A person re-identification feature extractor is incorporated using TriNet (Yuan et al. 2020) which uses a ResNet backbone to output descriptors that capture variations in human appearance. Re-identification of people can improve person identification during occlusions (Pinto et al. 2023). STARK was selected as a SOTA method due to its robustness to target occlusions and person appearance changes, while being computationally lightweight for embedded systems (Pinto et al. 2023). Moreover, it achieved the highest tracking results for tracking of partially occluded people in small crowds in indoor office environments on the KTP datasets.

In addition to RGB methods, we compared our tracking performance results to people tracking methods which use additional depth information to investigate whether depth has a significant influence as an additional mode. Specifically, we compare with available reported results for the KTP dataset, *e.g.*, Munaro *et al.* and Sun *et al.* (Munaro and Menegatti 2014; S. Sun et al. 2019).

**4. Munaro *et al.*** (Munaro and Menegatti 2014): This method uses RGB-D data with depth-based sub-clustering and HOG detection for tracking people using an Unscented KF (UKF) for position and velocity estimation and the Mahalanobis distance for data association. We choose this method as the use of a constant velocity model in the UFK allows for continuous tracking under temporary occlusions.

**5. Sun *et al.*** (S. Sun et al. 2019): RGB-D data is used within a Deformable Parts Model (DPM) for person detection and a Gaussian model is used to identify partially occluded people. A KF is used motion prediction and the Mahalanobis distance is used for data association. The method is able to effectively track individuals in scenarios with partial occlusions by evaluating depth values against a predefined Gaussian distribution to distinguish between occluded and visible parts of the target.

### 6.1.3 People Tracking Comparison Results

The people tracking results for our proposed *LDTrack* method and all comparison methods are presented in Table 1. In general, *LDTrack* outperformed the other RGB and RGB-D SOTA for people tracking by mobile robots with respect to both MOTA and MOTP across all test sets. This included under partial occlusions (KTP, ISRT, R-CPC datasets), pose deformations (ISRT, R-CPC datasets), and varying lighting (IOD, R-CPC datasets).

In particular, *LDTrack* achieved a MOTA of: 1) 78.6 on the IOD dataset, which is a 37% to 146% improvement over the other methods; 2) 92.7 on the KTP dataset, a 5% to 62% improvement; 3) 80.9 on the ISRT dataset, a 17% to 113% improvement; and 4) 49.0, 43.9, and 57.7 on Test Set 1, 2, and 3 from the R-CPC dataset with a 28% to 89% improvement over the SOTA methods. These results are statistically significant.

This shows that *LDTrack* can accurately identify and track individuals, distinguishing between different individuals as well as adapting to dynamic changes in person appearances in crowded and cluttered environments.

In general, the performance benefits of our *LDTrack* are due to its track-conditioned latent diffusion model, which dynamically updates person track embeddings by incorporating temporal information. This allows the model to adapt to changes in person appearances in the RGB images due to partial occlusions, and varying poses and lighting conditions. In contrast, the SOTA methods, which all use TBD approaches, rely on static person feature embeddings obtained from the detector at each frame. Such static embeddings can fail to account for temporal continuity, and are unable to adapt to changes in person appearances over time. For example, in scenarios where an individual transitions from sitting to standing, or becomes increasingly occluded.

In crowded environments in the R-CPC dataset, our *LDTrack* method outperformed the other methods with respect to both MOTA and MOTP. This demonstrates the advantage of our method in generating person box embeddings that can be iteratively refined to adapt to configurations in environments of any density or complexity, thereby improving generalizability. The REGROUP method showed improved performance over the other comparison methods with respect to MOTA and MOTP across all test sets on the R-CPC Dataset, due to its use of the CIF as previously mentioned. However, our *LDTrack* method still outperformed this method across all metrics, demonstrating the effectiveness of our diffusion model approach.

While *LDTrack* obtained the highest tracking performance on the R-CPC dataset, it performed worse on Test Sets 1 and 2, which contain corridors and hallways. These confined environments lead to frequent occlusions as individuals enter and exit the robot's field of view, often for only a single frame before reappearing many frames later, resulting in identity switches.

Non-parametric Kruskal-Wallis tests were performed on all datasets, showing a statistically significant difference in MOTA and MOTP between all methods: MOTA: $H = 31.5$, $p < 0.001$; and MOTP: ($H = 42.5$, $p < 0.001$). Posthoc Dunn tests with Bonferroni correction applied showed *LDTrack* had both a statistically significant higher 1) MOTA, $p < 0.001$, with respect to SORT ($Z = 3.03$), DeepSORT $Z = 3.92$), STARK ($Z = 4.08$), and REGROUP ($Z = 5.27$); and 2) MOTP, $p < 0.001$, with respect to SORT ($Z = 4.64$), DeepSORT ($Z = 4.79$), STARK ($Z = 5.53$), and REGROUP ($Z = 5.37$), respectively.

The *Munaro et al.* and *Sun et al* RGB-D methods, outperformed the SOTA *RGB-only* methods with respect to both MOTA and MOTP on the KTP dataset. This is due to the depth information providing additional spatial information to help distinguish between overlapping people and objects in occluded settings. However, despite being an RGB-only method, our *LDTrack* method still outperformed these two RGB-D methods across all metrics due to the inclusion of person track embeddings in our architecture. As these embeddings are specific to each individual, they allow for effective differentiation and re-identification of people using only RGB data.

Fig. 6 presents example tracking results of our *LDTrack* method (row 1) compared to the SORT (row 2), DeepSORT (row 3), STARK (row 4) and REGROUP (row 5) methods under: 1) poor illumination on the IOD dataset, Fig. 6(a), 2) partial occlusions by people on the KTP dataset, Fig. 6(b), 3) different poses and partial occlusions by both people and objects on the ISRT dataset, Fig. 6(c), and 4) varying poses and partial occlusions due to crowds on the R-CPC dataset, Fig. 6(d).

Fig. 6(a) represents a low-lighting scenario with two people walking in a hallway. The SOTA methods were unable to maintain consistent tracking, resulting in frequent loss of one or both people. In contrast, our *LDTrack* method was able to track the two people across all the consecutive frames. A scenario under partial occlusions is shown in Fig. 6(b) with two people on the left side of the images, one of whom is partially occluded by the other. There is an incorrect swap of the identity from the individual on the right to the one on the left in all the SOTA methods. Namely, the SORT and REGROUP methods, which do not consider appearance features such as clothing color and texture during data association, often resulted in identity swaps. While the DeepSORT and STARK methods do consider appearance features, they were unable to distinguish between the two individuals on the left due to similarities in their clothing. However, our *LDTrack* method was able to track both people consistently as distinct individuals since it explicitly considers temporal embedding information.

Fig. 6(c) includes a scene with people in sitting and leaning positions with occlusion by furniture/equipment or other people. The SORT, DeepSORT, and STARK methods only detected the upper portion of the foremost person, who was partially occluded by an experimental set-up. Unfortunately, the lower portion of the person was not detected in the bounding box. The REGROUP method incorrectly detected the upper and lower portions of this foremost person as separate people. However, our *LDTrack* was able to infer the continuity of the foremost partially occluded person as represented by one bound box. Fig. 6(d) presents a crowded environment where only the *LDTrack* method was able to successfully track all 9 individuals present, including the fourth person in the forefront from the left (*i.e.*, defined by the lavender bounding box), who varies their pose from bending over the table (away from the camera, frames 1-3), to standing (frame 4). In contrast, the other methods misidentify this person in different poses as separate people (by assigning different identities) across the frames.

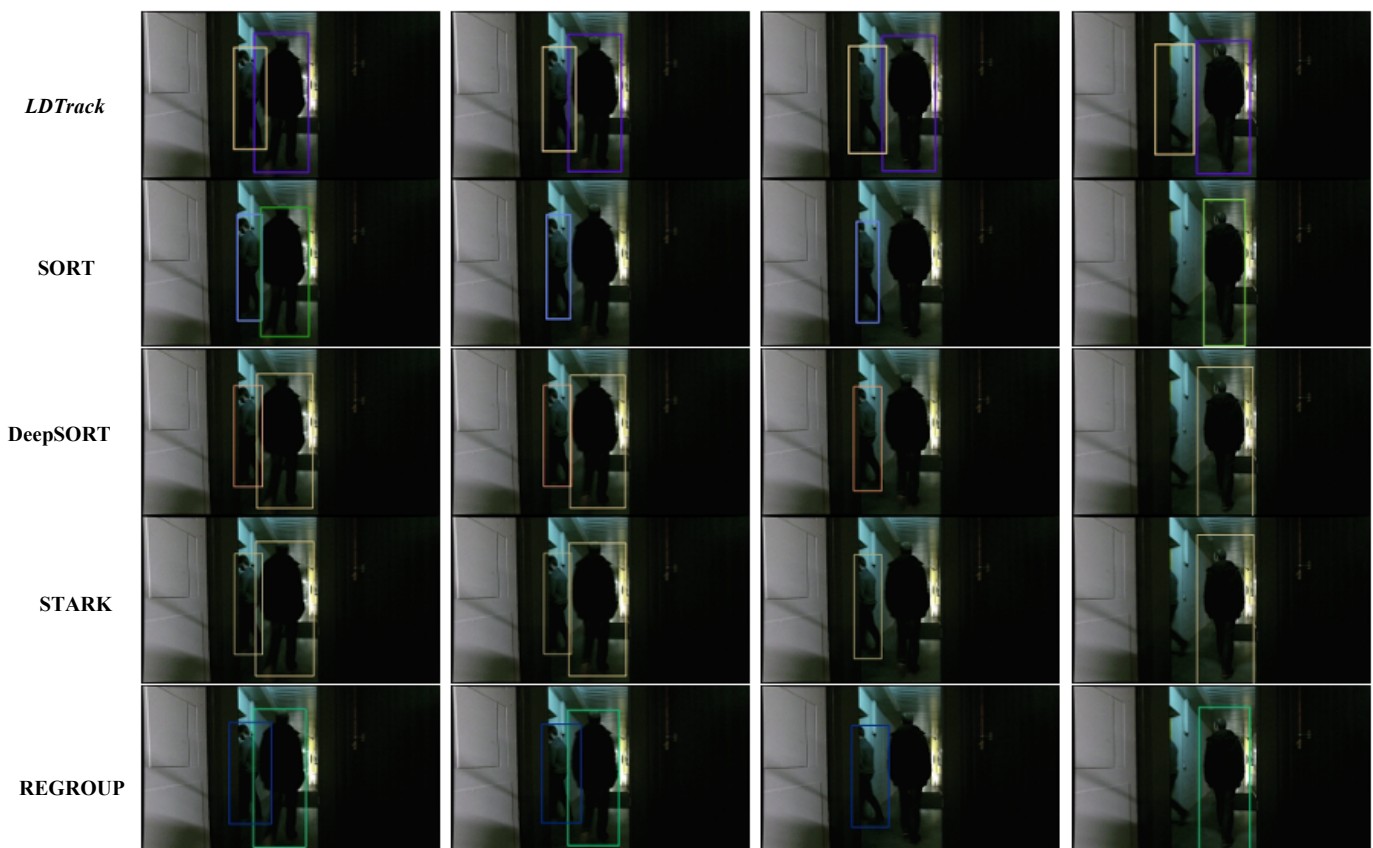

**(a)** IOD Sequence

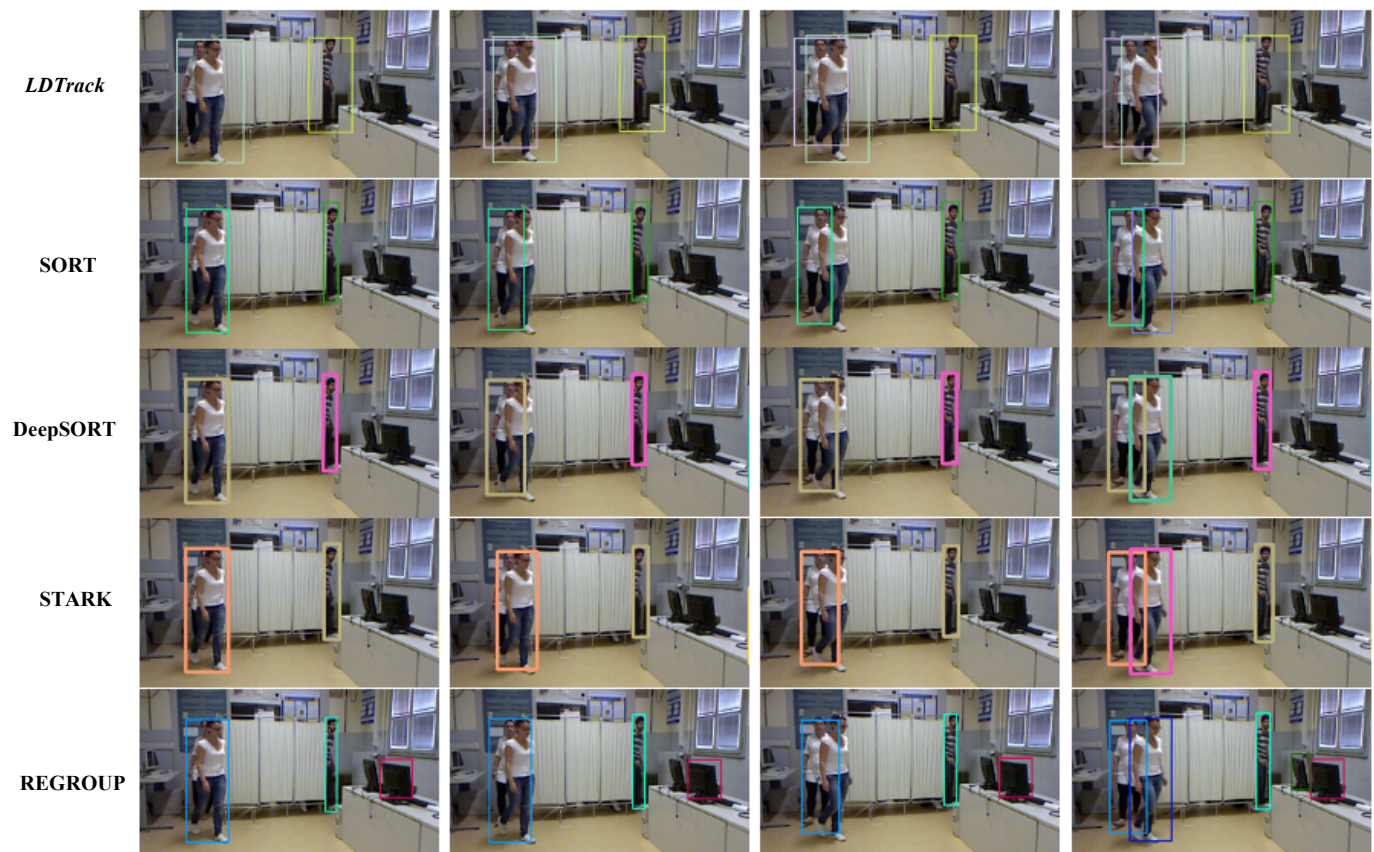

**(b)** KTP Sequence

**Fig. 6** People tracking results for *LDTrack*, SORT, DeepSORT, STARK and REGROUP: **(a)** under poor illumination, **(b)** partial occlusion

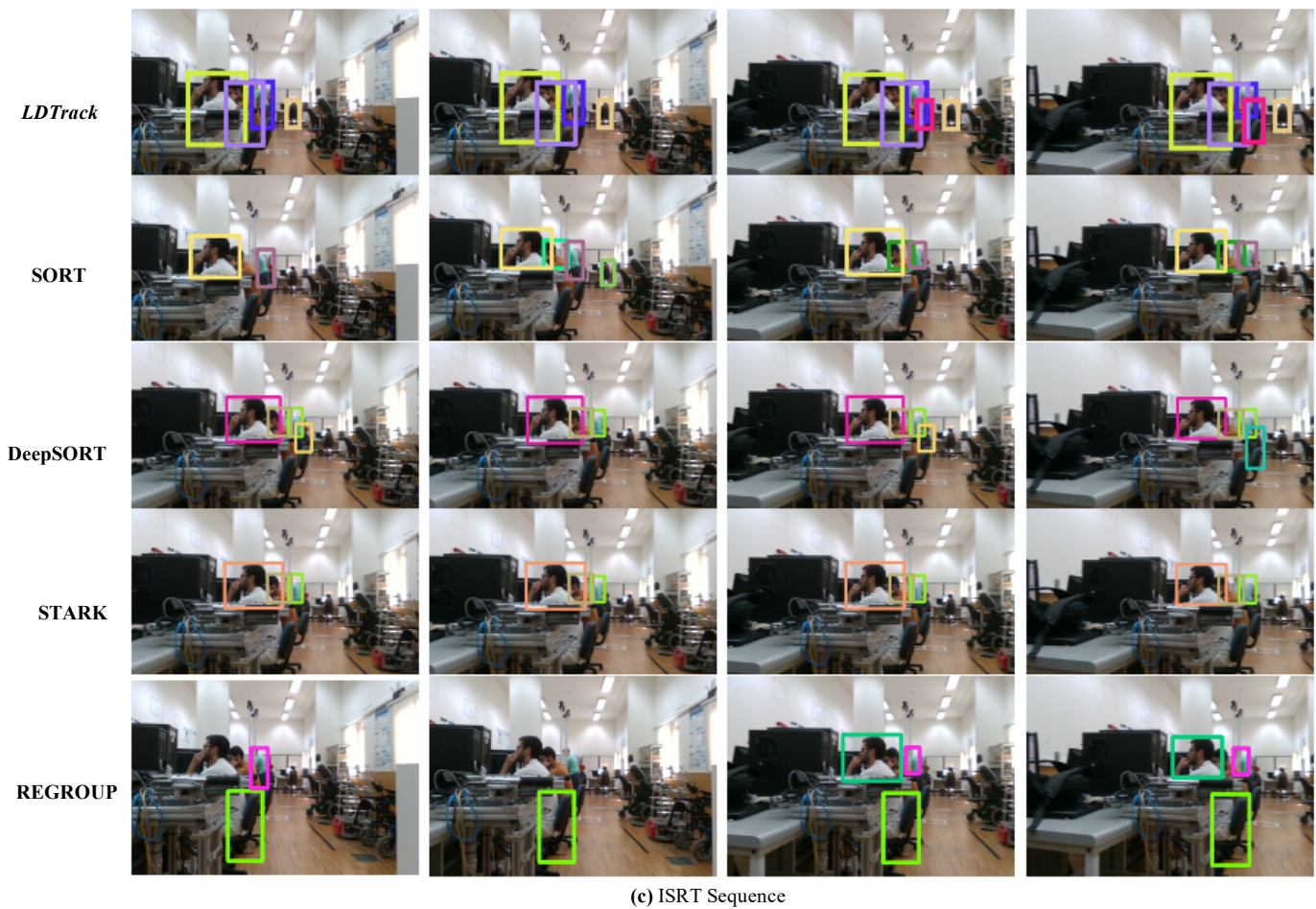

**(c)** ISRT Sequence

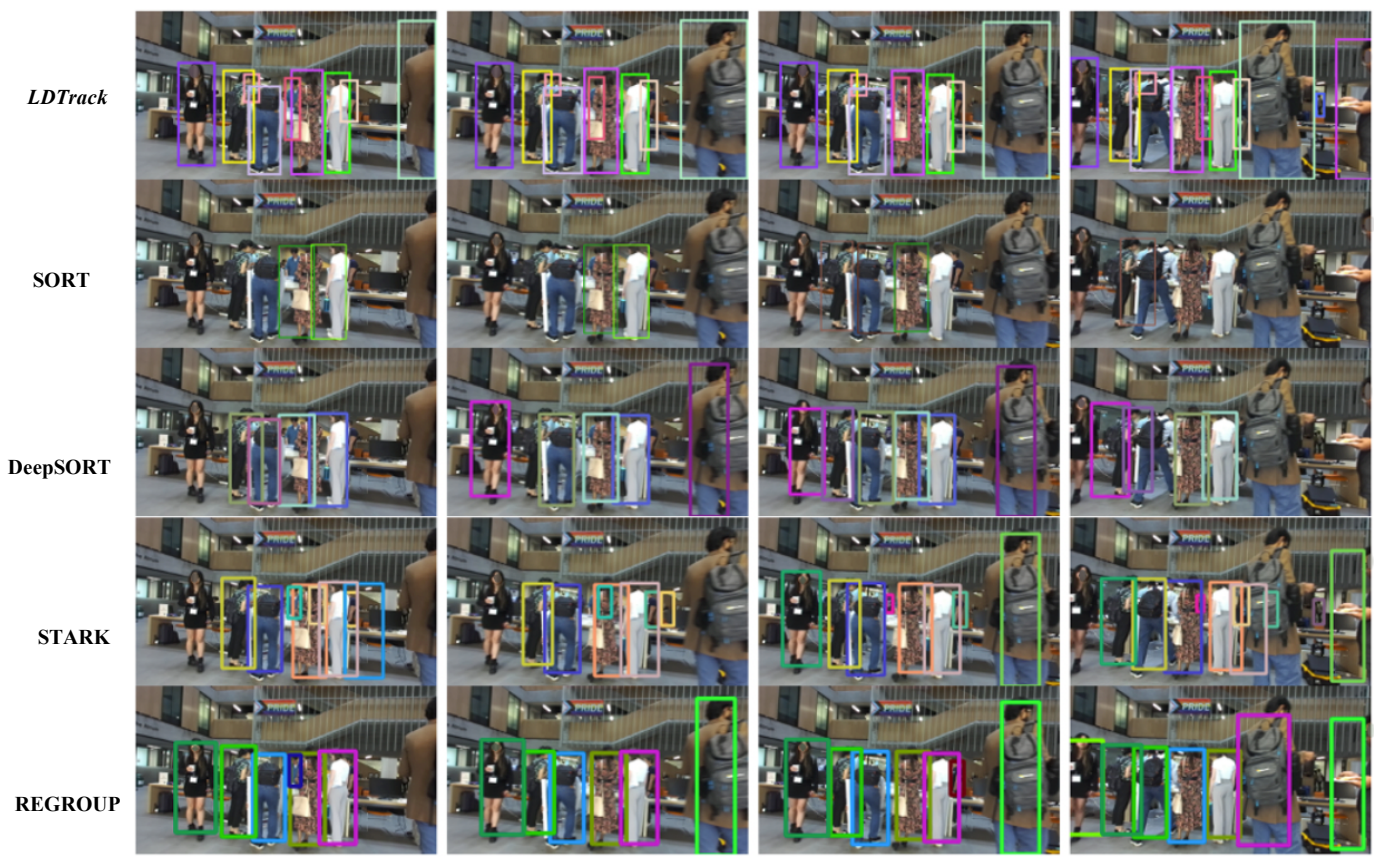

**(d)** R-CPC sequence

**Fig. 6** (continued) People tracking results for *LDTrack*, SORT, DeepSORT, STARK and REGROUP: **(c)** with different poses and partial occlusions, and **(d)** varying poses and partial occlusions due to crowds.

**Table 2** Comparison of MOTP* of our Proposed Tracking Method versus Existing Tracking Methods

| Method | IOD Test Set Lighting MOTP* | KTP Test Set Occ. MOTP* | ISRT Test Set Occ./Pose MOTP* | R-CPC Test Set 1 All MOTP* | R-CPC Test Set 2 All MOTP* | R-CPC Test Set 3 All MOTP* |
|---|---|---|---|---|---|---|
| *LDTrack* | 77.5 | 91.7 | 69.2 | 58.5 | 52.6 | 54.8 |
| SORT | 59.6 | 58.0 | 65.5 | 30.1 | 23.2 | 34.2 |
| DeepSORT | 62.9 | 55.4 | 65.3 | 38.2 | 31.4 | 38.8 |
| STARK | 62.6 | 55.9 | 65.4 | 40.9 | 34.4 | 39.9 |
| REGROUP | 61.1 | 55.2 | 56.3 | 40.0 | 35.2 | 36.5 |

We conducted a further analysis by comparing our *LDTrack* method with the SOTA methods using a modified version of MOTP, which we identify as MOTP*. We considered MOTP* as it explicitly considers missed detections, whereas MOTP only measures the precision of detected individuals, Eq. (14). This is particularly important in cluttered and crowded environments where missed detections due to occlusions can frequently occur. MOTP* is represented as:

$$\sum_{i=1}^{N_{match}} \text{IoU}\left(\boldsymbol{b}_i, \widehat{\boldsymbol{b}_i}\right)/N_{\text{GT}}, \qquad (16)$$

where IoU = 0 is assigned for each missed detection.

Applying MOTP* to all methods on the aforementioned IOD, KTP, ISRT, and R-CPC datasets in scenarios with person occlusions in dense crowds, significant improvements of *LDTrack* over the SOTA methods was observed, Table 2. In particular, on the R-CPC dataset, *LDTrack* obtained a MOTP* of 58.5 on Test Set 1, whereas SORT, DeepSORT, STARK, and REGROUP obtained a MOTP* of 30.1, 38.2, 40.9, and 40.0, respectively. Namely, there was a 44% to 94% improvement for *LDTrack* over the other methods. A non-parametric Kruskal-Wallis test was performed on all datasets, showing a statistically significant difference in MOTP* between all methods ($H = 28.9$, $p < 0.001$). Posthoc Dunn tests with Bonferroni correction applied showed our method had a statistically significant higher MOTP*, $p < 0.001$, with respect to SORT ($Z = 3.89$), DeepSORT ($Z = 3.68$), STARK ($Z = 4.32$) and REGROUP ($Z = 4.72$), respectively.

## 6.2 Multi-Object Tracking Comparison Study

We conduct a comparison study to evaluate the performance of *LDTrack* against multi-object tracking methods using the MOT17 and MOT20 datasets, which contain diverse urban environments with varying levels of pedestrian density, occlusions, and lighting conditions. For a fair comparison, we used a ResNet-50 backbone instead of ResNet-18 in the *SFEN* module of *LDTrack* to match the network capacities commonly used in such SOTA methods.

### 6.2.1 Performance Metrics

We used the following evaluation metrics for assessing object

tracking performance. Specifically:

1. **MOTA**

2. **IDF1**: This measures the ratio of correctly identified detections over the average number of ground truth and computed detections.

3. **IDSW**: This measures the number of instances a tracked trajectory changes its matched identity.

### 6.2.2 Comparison Methods

We compared *LDTrack* with SOTA joint detection and tracking methods such as: 1) TransTrack (Peize Sun et al. 2021), 2) Trackformer (Meinhardt et al. 2022), 3) MeMOT (Cai et al. 2022), 4) MOTR (Zeng et al. 2022), 5) CenterTrack (Zhou et al. 2020), 6) TraDes (Wu et al. 2021), 7) PermaTrack (Tokmakov et al. 2021), 8) TransCenter (Xu et al. 2022), 9) DiffusionTrack (Luo et al. 2023), 10) ConsistencyTrack (Jiang et al. 2024), and 11) MOTIP (Gao et al. 2024).

### 6.2.3 Datasets

We trained and evaluated on the following datasets:

1. **MOT17 (Dendorfer et al. 2021)**: This dataset consists of 14 video sequences captured in diverse urban environments, including shopping malls, pedestrian streets, and public squares. MOT17 presents scenarios with different illumination conditions due to the time-of-day and varying levels of occlusion.

2. **MOT20 (Voigtlaender et al. 2019)**: This dataset consists of 8 video sequences captured in densely crowded urban environments, including train stations, pedestrian streets, and public squares. MOT20 presents significant occlusions due to crowds and different illumination conditions due to night and daytime scenes.

### 6.2.4 MOT Comparison Results

The multi-object tracking results for our proposed *LDTrack* method and all comparison methods are presented in Table 3. This comparison evaluates the efficacy of diffusion models for tracking pedestrians in urban environments, while additionally demonstrating the generalizability of our method, which was designed for mobile service robots in human-centered environments, for urban environments. In general, our *LDTrack*, outperformed the SOTA methods with respect to MOTA on both the MOT17 and MOT20 datasets with up to 12.5% improvement on the MOT17 dataset, and up to 8.2% improvement on the MOT20 dataset. However, it has a slightly lower MOTA than Diffusion-Track on the MOT17 dataset. We postulate that this is due to DiffusionTrack incorporating both the current and previous RGB images, which doubles its input data. This leads to increasing the feature maps to twice their size, resulting in approximately double the memory usage. This increase in memory usage results in higher power consumption and computational costs due to the additional data being processed and stored. On the other hand, *LDTrack* only incorporates the current RGB image, resulting in a significantly lower memory footprint. This reduced memory footprint is advantageous for real-time applications in embedded mobile robot systems. Overall, *LDTrack* was able to generalize to tracking pedestrians in urban environments, despite not being specifically designed for multi-object tracking.

**Table 3** MOT Comparison Study with State-of-the-art Multi-Object Tracking Methods

| Dataset | MOT17 | | | MOT20 | | |
|---|---|---|---|---|---|---|
| | Test | | | Test | | |
| Method | MOTA | IDF1 | IDSW | MOTA | IDF1 | IDSW |
| *LDTrack* | 76.3 | 72.6 | 3795 | 68.9 | 64.5 | 2976 |
| *TransTrack* | 74.5 | 63.9 | 3663 | 64.5 | 59.2 | 3565 |
| *Trackformer* | 74.1 | 68.0 | 2829 | 68.6 | 65.7 | 1938 |
| *MeMOT* | 72.5 | 69.0 | 2724 | 63.7 | 66.1 | 1938 |
| *MOTR* | 71.9 | 68.4 | 2115 | - | - | - |
| *CenterTrack* | 67.8 | 64.7 | 3039 | - | - | - |
| *TraDes* | 69.1 | 63.9 | 3555 | - | - | - |
| *PermaTrack* | 73.8 | 68.9 | 3699 | - | - | - |
| *TransCenter* | 73.2 | 62.2 | 4614 | 67.7 | 58.7 | 3759 |
| *DiffusionTrack* | 77.9 | 73.8 | 3819 | 72.8 | 66.3 | 4117 |
| *ConsistencyTrack* | 69.9 | 65.7 | 3774 | - | - | - |
| *MOTIP* | 75.5 | 71.2 | - | - | - | - |

## 6.3 Ablation Study

We also performed an ablation study to evaluate the person track embedding design choice for the *LDTrack* architecture. Namely, we investigated three components:

1) Number of previous timesteps $N_{ts}$: We vary the number of past timesteps from which we retrieve person track embeddings $\boldsymbol{Z}_T^{s_0}$ as inputs to the *ITRN* module during inference had a direct effect on accuracy and precision. We varied the number of person track embeddings used as inputs from $N_{ts} = 1$ to $N_{ts} = 5$. For $N_{ts} = 1$, which is our current approach, we used only $\boldsymbol{Z}_{T-1}^{s_0}$ as input, which is the person track embeddings from the last timestep $T-1$; for $N_{ts} = 2$, we used both $\boldsymbol{Z}_{T-1}^{s_0}$ and $\boldsymbol{Z}_{T-2}^{s_0}$, following this procedure until $N = 5$, where we included all embeddings from $\boldsymbol{Z}_{T-1}^{s_0}$ to $\boldsymbol{Z}_{T-5}^{s_0}$ as inputs. Each person track, with identity $k$, is associated with multiple embeddings $\{\boldsymbol{Z}_{T-i}^{s_0}\}$, $i \in \{1,2,\dots,N_{ts}\}$ with each embedding $\boldsymbol{Z}_{T-i}^{s_0}$ generating a person bounding box prediction $\hat{b}_{k,T-i}$ with classification score $p(\hat{c}_{k,i-1})$. We selected the embedding with the highest classification score as the final person bounding box prediction, *e.g.*, $\hat{b}_{k,T_{\text{final}}} = \max_i \{\boldsymbol{Z}_{T-i}^{s_0}\}$.

2) Number of person box embeddings $N_{bx}$: We vary the number of person box embeddings generated by the *LBD* module to investigate the impact of increasing the number of proposals on tracking performance, using values of 250, 500, and 1000.

3) Latent space dimension $D$: We vary the latent space dimension of the *LFEN* module to investigate how the size of the person embeddings influences tracking performance, using values of 144, 288, and 566. In general, larger embedding sizes enable the learning of richer person feature representations at the cost of increased computational cost.

The results are presented in Table 3. Table 3 shows that increasing the number of person track embeddings from previous timesteps had little impact on both MOTA and MOTP. This can be attributed to the fact that these embeddings are continuously adapted and regressed over time through our *ITRN* module. Therefore, the person track embeddings from the most recent timestep $T-1$ incorporates all the relevant information regarding a person's appearance.

Regarding the number of person box embeddings $N_{bx}$ used during training, we observed that MOTA and MOTP improved significantly when increasing the number of embeddings $N_{bx}$ from 250 to 500, with diminishing returns observed from $N_{bx} = 500$ to $N_{bx} = 1000$. As each increase approximately doubles the computational cost, we selected $N_{bx} = 500$ as a trade-off between computational efficiency and tracking performance. Regarding the latent space dimension $D$, we observed a similar pattern. There was a performance gain when increasing the dimension from $D = 144$ to $D = 288$, however, there was only a slight improvement in MOTA and MOTP observed when increasing to $D = 566$. The ablation study validated our choice of using $N_{ts} = 1$, $N_{bx} = 500$, and $D = 288$, considering the trade-off between tracking performance and efficiency for real-time applications with mobile robots.

**Table 4** Ablation Study

| Dataset | IOD | | KTP | | ISRT | | R-CPC | | | | | |
|---|---|---|---|---|---|---|---|---|---|---|---|---|
| | Test Set Lighting | | Test Set Occlusion | | Test Set Occlusion/Pose | | Test Set 1 All | | Test Set 2 All | | Test Set 3 All | |
| Method | MOTP | MOTA | MOTP | MOTA | MOTP | MOTA | MOTP | MOTA | MOTP | MOTA | MOTP | MOTA |
| *Varying $N_{ts}$* | | | | | | | | | | | | |
| *LDTrack ($N_{ts} = 1$)* | 83.1 | 79.7 | 93.1 | 92.7 | 82.8 | 80.9 | 83.9 | 49.0 | 80.6 | 43.9 | 81.6 | 57.7 |
| *LDTrack ($N_{ts} = 2$)* | 83.1 | 79.7 | 92.4 | 92.0 | 82.9 | 80.5 | 84.0 | 49.1 | 80.9 | 42.4 | 81.9 | 56.2 |
| *LDTrack ($N_{ts} = 3$)* | 83.0 | 79.1 | 92.5 | 91.9 | 82.9 | 80.7 | 84.0 | 48.7 | 81.0 | 44.7 | 82.0 | 55.8 |
| *LDTrack ($N_{ts} = 4$)* | 83.0 | 79.1 | 92.5 | 91.9 | 82.8 | 81.3 | 84.1 | 48.7 | 81.0 | 42.3 | 81.9 | 56.4 |
| *LDTrack ($N_{ts} = 5$)* | 83.0 | 79.2 | 92.5 | 91.7 | 82.9 | 81.1 | 84.2 | 48.8 | 81.1 | 43.2 | 81.9 | 56.8 |
| *Varying $N_{bx}$* | | | | | | | | | | | | |
| *LDTrack ($N_{bx} = 250$)* | 81.4 | 78.2 | 91.5 | 91.1 | 81.4 | 79.8 | 82.3 | 47.2 | 78.6 | 42.4 | 80.2 | 56.1 |
| *LDTrack ($N_{bx} = 500$)* | 83.1 | 79.7 | 93.1 | 92.7 | 82.8 | 80.9 | 83.9 | 49.0 | 80.6 | 43.9 | 81.6 | 57.7 |
| *LDTrack ($N_{bx} = 1000$)* | 83.3 | 79.9 | 93.1 | 92.8 | 82.9 | 81.1 | 84.0 | 49.0 | 80.7 | 43.8 | 81.7 | 57.9 |
| *Varying $D$* | | | | | | | | | | | | |
| *LDTrack ($D = 144$)* | 80.8 | 76.7 | 91.0 | 91.1 | 78.5 | 78.1 | 80.3 | 46.6 | 76.8 | 41.2 | 77.7 | 54.9 |
| *LDTrack ($D = 288$)* | 83.1 | 79.7 | 93.1 | 92.7 | 82.8 | 80.9 | 83.9 | 49.0 | 80.6 | 43.9 | 81.6 | 57.7 |
| *LDTrack ($D = 566$)* | 83.5 | 80.1 | 93.2 | 92.9 | 83.1 | 81.3 | 84.3 | 49.3 | 81.0 | 44.2 | 82.1 | 58.1 |

## 7 Conclusion

In this paper, we present a novel mobile robot person-tracking architecture, *LDTrack*, which directly addresses intraclass variations through the unique use of conditional latent diffusion models. The diffusion process is implemented in latent space, enabling the spatial-temporal refinement of temporal person embeddings. These embeddings represent different individuals detected by a robot in a shared environment, adapting dynamically to changes in their appearance over time. Such adaptation is essential in cluttered and crowded environments, where the appearance of individuals can vary significantly over time due to factors such as occlusions, pose deformations, and lighting variations. This adaptation is achieved by conditioning the diffusion model on person track embeddings from the previous timestep. Extensive experiments verified that our *LDTrack* outperforms existing state-of-the-art tracking methods for mobile robots in tracking dynamic people across multiple real-world crowded and cluttered environments, as well as for tracking pedestrians in outdoor urban environments. An ablation study further validated our design choices of *LDTrack*. Future work will explore the integration of our previously developed contrastive learning approach TimCLR (Fung et al. 2023), to pretrain our SFEN network in order to learn person representations that are invariant under intraclass variations. We will also test this combined methodology in real-time on a mobile robot in cluttered environments.

## 8 Declarations

### 8.1 Funding

This work was supported by the Natural Sciences and Engineering Research Council of Canada (NSERC), AGE-WELL Inc., and the Canada Research Chairs (CRC) program.

### 8.2 Conflicts of interest/Competing interests

We have no known conflicts of interest.

### 8.3 Availability of Data and Material

The IOD, KTP, ISRT, MOT17, and MOT20 datasets are available at:
1) IOD (Mees et al. 2016)
2) KTP (Munaro and Menegatti 2014)
3) ISRT (Pereira et al. 2022)
4) MOT17 (Dendorfer et al. 2021)
5) MOT20 (Voigtlaender et al. 2019)

## 8 Acknowledgements

The authors would like to thank Aaron Hao Tan and Haitong Wang for their invaluable discussions and assistance.